\documentclass{article}

    \usepackage[final, nonatbib]{neurips_2020}

\usepackage[utf8]{inputenc} 
\usepackage[T1]{fontenc}    
\usepackage{hyperref}       
\usepackage{url}            
\usepackage{booktabs}       
\usepackage{amsfonts}       
\usepackage{nicefrac}       
\usepackage{microtype}      

\usepackage{graphicx}
\graphicspath{ {./images/} }
\usepackage[utf8]{inputenc}
\usepackage{authblk}

\title{A Framework and Dataset for Abstract Art Generation via CalligraphyGAN}

\author[1]{Jinggang Zhuo}
\author[1]{Ling Fan}
\author[2]{Harry Jiannan Wang}
\affil[1]{Design AI Lab, Tongji University, Shanghai, China}
\affil[2]{
    University of Delaware, Newark, DE, USA\authorcr
    \texttt{\{zhuojg1519, lfan\}@tongji.edu.cn, hjwang@udel.edu}
}

\begin{document}

\maketitle

\section{Introduction}

With the advancement of deep learning, artificial intelligence (AI) has made many breakthroughs in recent years and achieved superhuman performance in various tasks such as object detection, reading comprehension, and video games. Generative Modeling such as various Generative Adversarial Networks (GAN)~\cite{goodfellow2014generative} models has been applied to generate paintings and music. Research in Natural Language Processing (NLP) also had a leap forward in 2018 since the release of the pre-trained contextual neural language models such as BERT~\cite{DBLP:journals/corr/abs-1810-04805} and recently released GPT3~\cite{brown2020language}. Despite the exciting AI applications aforementioned, AI is still significantly lagging behind humans in creativity, which is often considered the ultimate moonshot for AI. Our work is inspired by Chinese calligraphy, which is a unique form of visual art where the character itself is an aesthetic painting. We also draw inspirations from paintings of the Abstract Expressionist movement in the 1940s and 1950s, such as the work by American painter Franz Kline \footnote{https://www.moma.org/collection/works/79234}. In this paper, we present a creative framework based on Conditional Generative Adversarial Networks~\cite{DBLP:journals/corr/MirzaO14} and Contextual Neural Language Model to generate abstract artworks that have intrinsic meaning and aesthetic value, which is different from the existing work, such as image captioning and text-to-image generation, where the texts are the descriptions of the images. In addition, we have publicly released a Chinese calligraphy image dataset \footnote{https://github.com/zhuojg/chinese-calligraphy-dataset} and demonstrate our framework using a prototype system and a user study.

\section{Research Framework and Dataset}

\begin{figure}
    \centering
    \includegraphics[width=\textwidth]{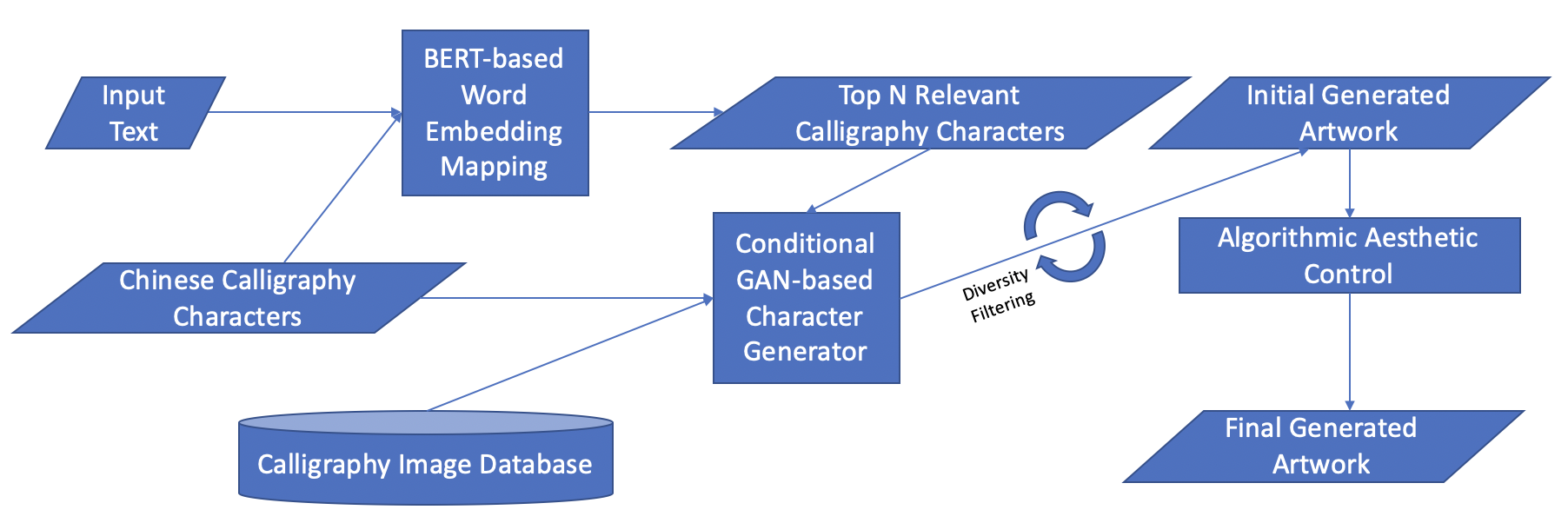}
    \caption{\label{fig:framework}Our proposed framework}
\end{figure}

Inspired by Pablo Picasso, art washes away from the soul and dust of everyday life. Our current target application is to generate abstract arts drawing from a most typical daily scene: we will use dish names and their images to generate abstract arts. Figure \ref{fig:framework} shows our proposed framework. We want to use Chinese calligraphy to link the dish name text with the final abstract art to enable its intrinsic meaning. We collected 138,499 images of Chinese calligraphy characters written by 19 calligraphers from the Internet, which cover 7328 different characters in total. We chose 1000 characters with more than 25 different images each to train a slightly revised conditional GAN~\cite{DBLP:journals/corr/MirzaO14} named CalligraphyGAN using each character as the control condition. Due to the dataset limitation, we developed a simple algorithm based on BERT~\cite{DBLP:journals/corr/abs-1810-04805} to map the input text, i.e., Chinese dish names, with arbitrary number of characters into five characters from the 1000 characters : 1) generate the embeddings for the input text and each of the 1000 characters using the pre-trained BERT model; 2) calculate the pairwise similarity between the input text embedding and character embedding; 3) choose the top 5 characters with the highest similarity scores. Then, we use the five chosen characters as the controlling conditions to generate a new character using CalligraphyGAN, which has never been seen before but embeds the shape characteristics from all five characters. Note that we use equal weights for each character but can adjust weights to draw more attention to any specific character. In the prototype system, we feed the dish names with some random noises so that the same dish can generate different artwork each time. Fréchet Inception Distance (FID)~\cite{heusel2017gans} is often used to measure the quality and diversity of generated images and we generate 50 images and choose the one with lowest FID score. Next, we conduct a number of algorithmic aesthetic control: 1) denoising the image to clean up the background; 2) applying an algorithm to transfer the image to oil-painting style using key colors automatically extracted from the dish image; 3) providing optional additional style transfer~\cite{ghiasi2017exploring} based on paintings from famous painters such as Picasso, Rothko, Pollock, and de Kooning. The last step is to generate the final artwork by combining multiple elements, such as the generated image, a piece of text description, a company logo, etc. with a smart layout algorithm. 

\section{Prototype and Evaluation}

\begin{figure}
    \centering
    \includegraphics[width=350pt]{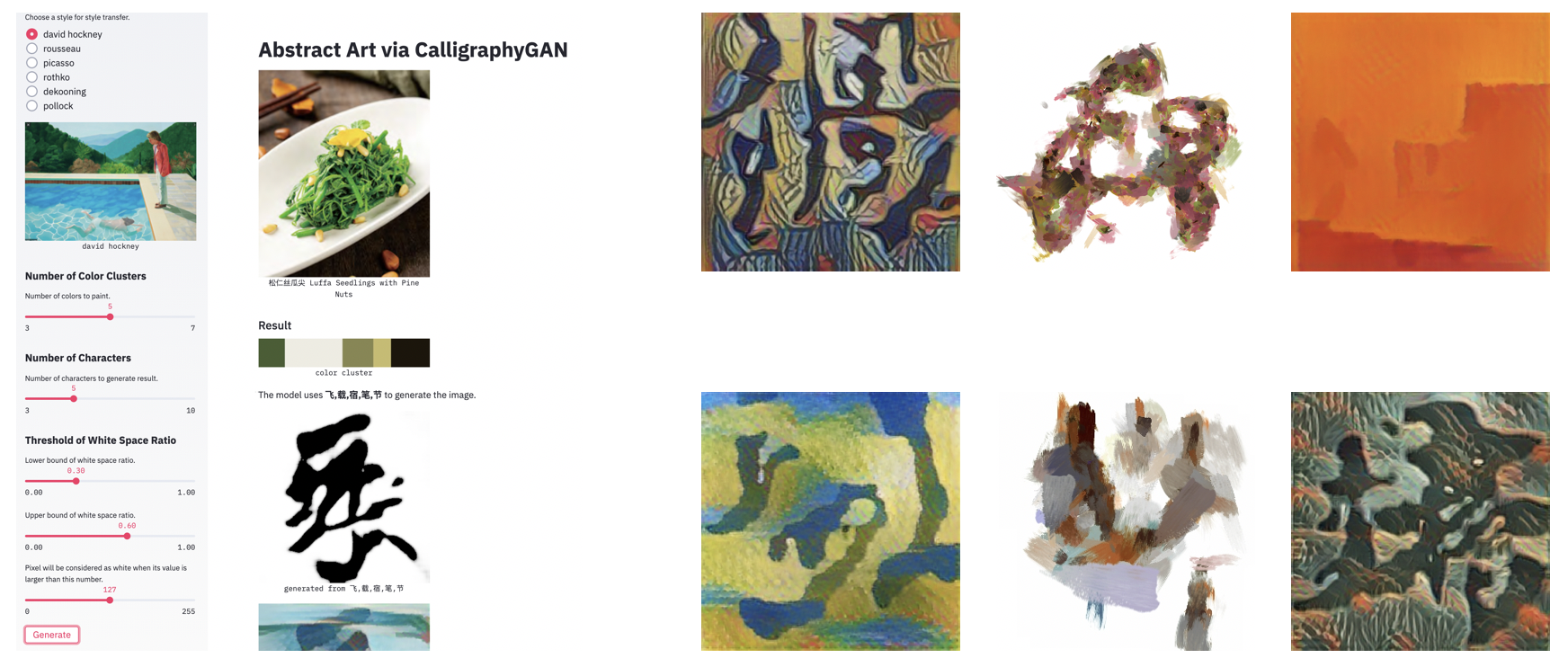}
    \caption{\label{fig:demo}Prototype System and Sample Generated Art}
\end{figure}

We partner with a restaurant in Shanghai, China to build a prototype system \footnote{http://harrywang.me/calligraphy/} to demonstrate our framework. The users can customize various parameters, such as style transfer sample, number of key colors extracted, and the ratio of white spaces, to control the generation. We asked 8 users to try the prototype systems using different dish samples for at least 8 times each and asked them to pick their favorite result from the generated arts. Figure \ref{fig:demo} shows the system UI and some favorite generation results. We are in the process of deploying a pilot system to project the generated artworks onto the table cloth via a gesture-controlled projector, using which the restaurant guests can interact with our artwork, such as generating a new artwork, choosing different style transfer options, and giving rating and feedback. To the best of our knowledge, this is the first of its kind of applying AI creativity in a restaurant setting to create a unique customer dining experience. 

\bibliographystyle{abbrv}
\bibliography{neurips_2020.bib}

\begin{thebibliography}{1}

\bibitem{brown2020language}
T.~B. Brown, B.~Mann, N.~Ryder, M.~Subbiah, J.~Kaplan, P.~Dhariwal,
  A.~Neelakantan, P.~Shyam, G.~Sastry, A.~Askell, et~al.
\newblock Language models are few-shot learners.
\newblock {\em arXiv preprint arXiv:2005.14165}, 2020.

\bibitem{DBLP:journals/corr/abs-1810-04805}
J.~Devlin, M.~Chang, K.~Lee, and K.~Toutanova.
\newblock {BERT:} pre-training of deep bidirectional transformers for language
  understanding.
\newblock {\em CoRR}, abs/1810.04805, 2018.

\bibitem{ghiasi2017exploring}
G.~Ghiasi, H.~Lee, M.~Kudlur, V.~Dumoulin, and J.~Shlens.
\newblock Exploring the structure of a real-time, arbitrary neural artistic
  stylization network.
\newblock {\em arXiv preprint arXiv:1705.06830}, 2017.

\bibitem{goodfellow2014generative}
I.~Goodfellow, J.~Pouget-Abadie, M.~Mirza, B.~Xu, D.~Warde-Farley, S.~Ozair,
  A.~Courville, and Y.~Bengio.
\newblock Generative adversarial nets.
\newblock In {\em Advances in neural information processing systems}, pages
  2672--2680, 2014.

\bibitem{heusel2017gans}
M.~Heusel, H.~Ramsauer, T.~Unterthiner, B.~Nessler, and S.~Hochreiter.
\newblock Gans trained by a two time-scale update rule converge to a local nash
  equilibrium.
\newblock In {\em Advances in neural information processing systems}, pages
  6626--6637, 2017.

\bibitem{DBLP:journals/corr/MirzaO14}
M.~Mirza and S.~Osindero.
\newblock Conditional generative adversarial nets.
\newblock {\em CoRR}, abs/1411.1784, 2014.

\end{thebibliography}

\end{document}